\documentclass[runningheads]{llncs}
\usepackage{amsmath,amssymb} 
\usepackage{color}
\usepackage[width=122mm,left=12mm,paperwidth=146mm,height=193mm,top=12mm,paperheight=217mm]{geometry}

\usepackage{float}
\usepackage{graphics}
\usepackage{algorithm}
\usepackage[noend]{algpseudocode}

\usepackage{tikz}

\begin{document}
\pagestyle{headings}
\mainmatter
\def\ECCV18SubNumber{1624}  

\title{SaaS: Speed as a Supervisor \\ for Semi-supervised Learning}
\titlerunning{SaaS: Speed as a Supervisor for Semi-supervised Learning}
\authorrunning{Safa Cicek, Alhussein Fawzi and Stefano Soatto}
\author{Safa Cicek, Alhussein Fawzi and Stefano Soatto}
\institute{University of California, Los Angeles\\
    \email{ \{safacicek,fawzi,soatto\}@ucla.edu}
}

\maketitle

\begin{abstract}
We introduce the SaaS Algorithm for semi-supervised learning, which uses learning speed during stochastic gradient descent in a deep neural network to measure the quality of an iterative estimate of the posterior probability of unknown labels. Training speed in supervised learning correlates strongly with the percentage of correct labels, so we use it as an inference criterion for the unknown labels, without attempting to infer the model parameters at first. Despite its simplicity, SaaS achieves state-of-the-art results in semi-supervised learning benchmarks.
\end{abstract}

\begin{figure}
  \centering
    \includegraphics[width=5cm]{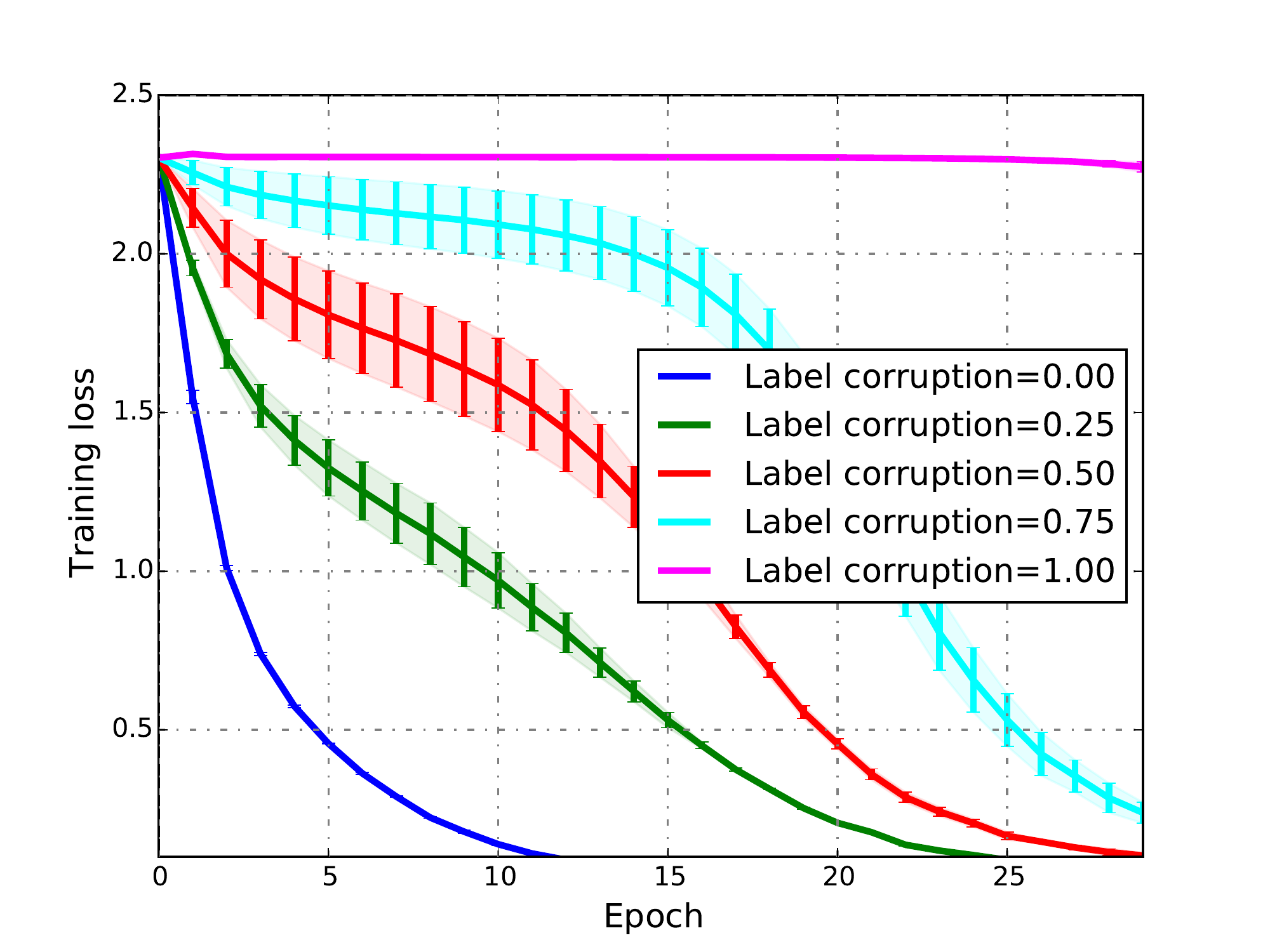}
	\includegraphics[width=5cm]{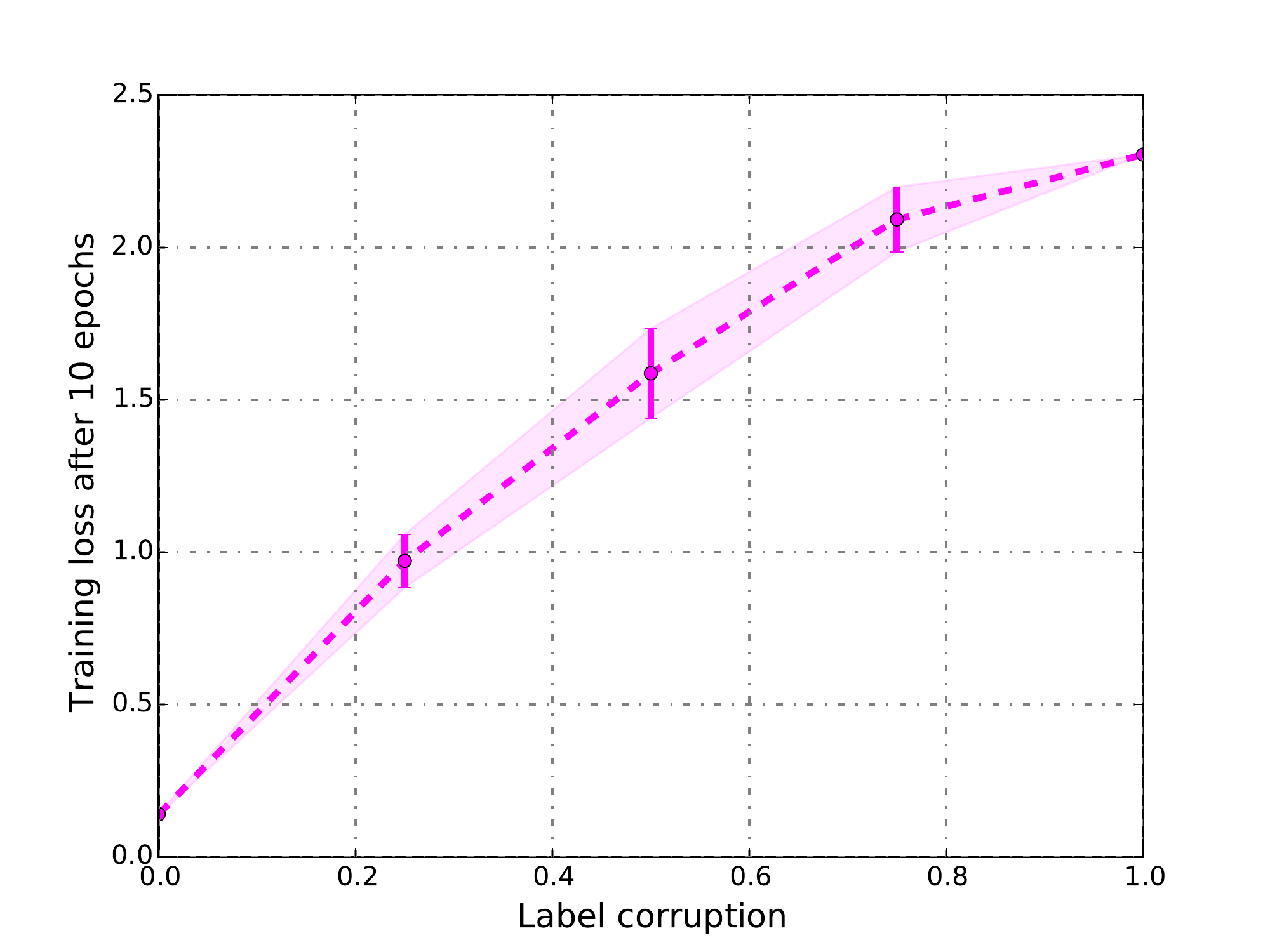}
\caption{ {\bf Supervision quality affects learning speed.} During training, the loss decreases rapidly when most labels provided are correct, and slows down significantly as the percentage of correct labels decreases. The left plot shows the loss when training a Resnet18 on CIFAR10 for different percentages of corrupted labels. The error bars show mean and standard deviation over 3 runs with random initial weights.
The right panel shows the loss as a function of the percentage of incorrect labels for a unit of time corresponding to ten epochs. 
All the results use a fixed learning rate of $0.1$, with no data augmentation or weight decay.}
\label{fig-random-label}
\end{figure}

\section{Introduction}

The key idea of our approach is to {\em use speed of convergence as an inference criterion for the value of the unknown labels} for semi-supervised learning (SSL). Fig. \ref{fig-random-label} explicitly shows the relation between label corruption and training speed.

In SSL, one is given some labeled and some unlabeled data to train (infer the parameters of) a classifier, in hope of it performing better than if trained on the labeled data alone \cite{chapelle2009semi}. This is an important problem in vision where annotations are costly but unlabeled data are aplenty. 

To measure learning speed, we use a small number of epochs as the unit of time, and compute the decrease of the loss in that interval when following a standard optimization procedure (stochastic gradient descent or Langevin dynamics). The main idea of our SSL algorithm is then to optimize the labels of unlabeled data (or more precisely, the posterior distribution of the unlabeled data) to maximize the loss decrease. 
The resulting SaaS Algorithm 
is composed of an \textit{outer loop}, which updates the distribution of unknown labels, and an \textit{inner loop} which simulates the optimization procedure over a small number of epochs.  
The proposed algorithm is unusual, as the posterior distribution of the unknown labels is initially inferred {\em independently of the model parameters} (weights), rather than along with them as customary in SSL. 
The training loss also includes explicit regularization, making it an Information Bottleneck Lagrangian,  consistent with recent theoretical developments in Deep Learning.

Despite its simplicity, SaaS achieves state-of-the-art results, reported in Sect. \ref{sect-empirical}. In the next section we formalize our method, and in Sect. \ref{sect-discussion} we discuss our contributions in relation to the prior art, and highlight its features and limitations.

\subsection{Description of the method}

We are given some labeled data ${\bf x}^l \doteq \{x_i^l\}_{i=1}^{N^l}$ with labels ${\bf y}^l \doteq \{y_i^l\}_{i=1}^{N^l}$ and some unlabeled data, ${\bf x}^u \doteq \{x_i^u\}_{i=1}^{N^u}$. The unknown labels ${\bf Y}^u \doteq \{Y_i^u\}_{i=1}^{N^u}$ are hidden variables whose ``true values'' $ y^u_i$ are not of interest {\em per se}, but must be dealt with (nuisance variables). Most SSL approaches attempt to infer or marginalize the unknown labels along with the model parameters, for us the weights $w$ of a neural network, only to discard the former and keep the weights.\footnote{In some cases, one can compute a modified functional that is invariant to nuisance variables, yet sufficient for the model parameters, for instance in separable least-squares problems.}

Unlike most SSL approaches, in our approach  we {\em estimate the posterior distribution of the unknown labels} $P^u_i \doteq P(Y^u_i | X^u_i = x^u_i)$. The outer loop of the algorithm updates the estimates of the posterior $P^u_i$, while the inner loop optimizes over the weights (for the fixed estimate of the posterior) to estimate the loss decrease over the time interval. It is important to note that we are \emph{not attempting to infer the weights} (but only the posterior distribution of the unknown labels), which  
are resampled at each (outer) iteration.\footnote{We have tested both drawing the weights from a Gaussian distribution, or resetting them to their initial value, which yields similar results.} By design, the weights do not converge, but empirically we observe that the posterior distribution of the unknown labels does. We then use the maximum a-posteriori estimate of the labels $\hat y_i^u = \arg\max_i P^u_i$  to infer a point-estimate of the weights $\hat w$ in a standard supervised training session. This procedure is described in the SaaS algorithm in Sect. 2, where $\ell$ is a loss function described in Sect.~\ref{sect-description}, $N = N^u + N^l$ is the total number of samples and $\eta$ are suitable learning rates for a batch size $|B|$ in SGD.

In Sect.~\ref{sect-empirical} we test the SaaS algorithm on SSL benchmarks, and in Sect.~\ref{sect-discussion} we place our contribution in the context of related literature. Next, we derive the algorithm in greater detail.


\section{Derivation of the model}
\label{sect-description}

We represent a deep neural network with parameters (weights) $w$, trained for classification into one of $K$ classes, as a function $f_w(x) \in \mathbb R^K$ where $x$ is the input (test) datum, and the $k$-th component of the output approximates the posterior probability $f_{w_t}(x_i)[k] \simeq P(y_i = k | x_i)$.
Stochastic gradient descent (SGD) performs incremental updates of the unknown parameters with each iteration $t$ computing the loss by summing on a random subset of the training set called ``mini-batch'' $B_t$. The number of iterations needed to sample all the dataset is called an epoch. We represent SGD as an operator $G(\cdot): w_t \rightarrow w_{t+1}$, which maps the current estimate of the weights to the next one. Note that $G$ depends on the the given (true) labels, and the hypothesized ones for the unlabeled data. 

To quantify \textit{learning speed} we use the cumulative loss in a fixed time (epoch) interval: For a given training set $\{{\bf x}, {\bf y} \}$, it is the aggregated loss during $T$ optimization steps, {\em i.e.}, the area under the learning curve \begin{align}
{\cal L}_{T} 
= \frac{1}{T} \sum_{t=1}^T \frac{1}{|B_t|} \sum_{i = 1}^{|B_t|} \ell(x_i, y_i; w_t)
\end{align} where $|B_t|$ denotes the cardinality of the mini-batch, composed of samples \\ $(x_i, y_i) \sim P(x, y)$; $\ell$ denotes the classification loss corresponding to weights $w_t$. Computing the loss above over all the data points requires the labels being known. Alternatively, it can be interpreted as a loss for a joint hypothesis for the weights $w_t$ {\em and} the label $y_i$. We use the cross-entropy loss, which is the sampled version of $H_{P, Q}(y | x) = {\mathbb E}_{P(x)}{\mathbb E}_{P(y|x)} -\log Q(y|x)$ where $Q(y = k|x) = f_w(x)[k]$ is the $k$th coordinate of the output of the network. 

The true joint distribution $P(x)P(y|x) =  P(x, y)$ is not known, but the dataset is sampled from it.  In particular, if $y_i = k$ is the true label for $x_i$, we have $P(y_i|x_i) = \delta(y_i - k)$. Otherwise, we represent it as an unknown $K$-dimensional probability vector $P^u_i$ with $k$-th component $P^u_i[k] = P(y_i = k | x_i), \ k = 1, \dots, K$, to be inferred along with the unknown weights $w$. We can write the sum \\ $\sum_{k = 1}^K P_i[k]P_j[k]$ as an inner product between the probability vectors $\langle P_i, P_j\rangle = P_i^T P_j$, so that the \textit{cumulative loss} can be written as 
\begin{equation}
\label{eq:L_TPu}
{\cal L}_T(P^u) =  \frac{1}{T} \sum_{t=1}^T \frac{1}{|B^u_t|} \sum_{i = 1}^{|B^u_t|} - \underbrace{\langle \log f_{w_t}(x^u_i),  P^u_i\rangle}_{\ell(x^u_i, P^u_i; w_t)}   
\end{equation} where $B^u_t$ is mini-batch of unlabeled samples at iteration $t$. Note that cross-entropy depends on the {\em posterior distribution} of the unknown labels, $P^u_i$, rather than their sample value $y^u_i$. The loss depends on the posterior for the entire unlabeled set, which we indicate as an $N^u \times K$ matrix $P^u$, and the entire set of weights $w = \{w_1, \dots, w_T\}$. 

We also add as an explicit regularizer the entropy of the network outputs for the unlabeled samples: $ -{\mathbb E}_Q \log Q(y^u | x^u)$, as common in SSL \cite{grandvalet2005semi}, which we approximate with the unlabeled samples as
\begin{equation}
H_Q(w) = \sum_{i=1}^{N^u} -\underbrace{\langle f_w(x_i^u), \log f_w(x_i^u)\rangle}_{q(x_i^u; w)}
\end{equation}
We further incorporate data augmentation by averaging over group transformations $g(x) \in {\mathbb G}$, such as translation and horizontal flipping, sampled uniformly $g_i \sim {\cal U}({\mathbb G})$. Let us define the following shorthand notations: $\ell(B^u_t, P^u; w_{t-1}) = \frac{1}{|B^u_t|} \sum_{i = 1}^{|B^u_t|} \ell(g_i(x^u_i), P^u_i; w_{t-1})$ and $q(B^u_t; w_{t-1})=\frac{1}{|B^u_t|} \sum_{i = 1}^{|B^u_t|} q(g_i(x_i^u); w_{t-1})$. Similarly for labeled set, $\ell(B^l_t, P^l; w_{t-1}) = \frac{1}{|B^l_t|} \sum_{i = 1}^{|B^l_t|} \ell(g_i(x^l_i), P^l_i; w_{t-1})$ where $P^l_i = \delta_{i,k}$ is the Kronecker Delta with $k$ the true label associated to $x^l_i$. The overall learning can be framed as the following optimization 
\begin{align}
\label{eq:obj}
P^u = \arg\min_{P^u}&  \frac{1}{T} \sum_{t=1}^T \ell(B^u_t, P^u; w_{t-1}) \\
{\rm subject \ to} &\ w_{t-\frac{1}{2}} = w_{t-1} - \eta_{w} \nabla_{w_{t-1}} \left( \ell(B^u_t, P^u; w_{t-1})+ \beta q(B^u_t; w_{t-1}) \right) \nonumber \\
&\ w_{t} = w_{t-\frac{1}{2}} - \eta_{w} \nabla_{w_{t-\frac{1}{2}}} \ell(B^l_t, P^l; w_{t-\frac{1}{2}}) \  \forall \ t = 1 \dots T \nonumber \\
&\ P^u \in \mathcal{S} \nonumber
\end{align} 

where the last constraint imposes that the columns of $P^u$ be in the probability simplex of $\mathbb{R}^K$. The objective of the above optimization is to find the posterior of the unlabeled data that leads to the fastest learning curve, when using stochastic gradient descent to train the weights $w$ on both labeled and unlabeled data. The update of the weights is specifically decomposed into two steps: the first step is an update equation for the weights with \textit{unlabeled} samples and posterior $P^u$, while the second step updates the weights using the labeled samples and ground truth labels. We stress that the latter update is crucial in order to fit the weights to the available training data, and hence prevents from learning trivial solutions of $P^u$ that lead to a fast convergence rate, but does not fit the data properly. We also note that the entropy term is only minimized for unlabeled samples. We set $\beta=1$ in all the experiments. 

It is customary to regularize the labels in SSL using entropy or a proxy \cite{miyato2017virtual,dai2017good,grandvalet2005semi,krause2010discriminative,springenberg2015unsupervised}, including mutual exclusivity \cite{sajjadi2016mutual,xu2017semantic}. \cite{shrivastava2012constrained} uses mutual exclusivity adaptively by not forcing it in the early epochs for similar categories. 
All these losses force decision boundaries to be in the low-density region, a desired property under cluster assumptions. \cite{krause2010discriminative,springenberg2015unsupervised} also maximize the entropy of the marginal label distribution to balance the classes. Together with entropy minimization, balancing classes is equivalent to maximizing the mutual information between estimates and the data if the label prior is uniform. However, we did not apply this loss to not restrict ourselves to balanced datasets or to the settings where we have prior knowledge on the label distributions. 


The $\beta$ factor modulates the entropy of the unknown labels, which acts as an explicit regularizer. The weights will be regularized implicitly by the learning algorithm since we use SGD that has been shown to have an inductive bias of similar form \cite{chaudhari2017stochastic}.
Together, these make the loss function above into an Information Bottleneck Lagrangian, \cite{achille2017emergence}. This also explains why our method would not work for general classifiers, as it requires the implicit bias of SGD for generalization.

\subsection{Implementation}


To solve the optimization problem in Eq. \eqref{eq:obj}, we perform gradient descent over the unknowns $P^u$, where $P^u$ is the  unknown-label posterior initialized randomly. Starting from $w_0$ sampled from a Gaussian distribution, the inner loop performs a few epochs of SGD to measure learning speed (cumulative loss) ${\cal L}_T$ \textit{while keeping the label posterior fixed}. The outer loop then applies a gradient step to update the unknown-label posterior $P^u$. After each update, the weights are either reset to $w_0$, or resampled from the Gaussian. In the beginning of each outer epoch, label estimates $P^u \in \mathbb R^{N^u \times K}$ are projected with operation $\Pi(P^u)$ to the closest point on the probability simplex of dimension $N^u \times K$.

After the label posterior converges (the weights never do, by design, in the first phase), we select the maximum a-posteriori estimate $\hat y^u_i = \arg\max_i P^u_i$, and proceed with training as if fully supervised in the second phase. We call the resulting algorithm, described in Algorithm \ref{pseudo}, SaaS.  

\begin{algorithm}
\begin{footnotesize}
\caption{SaaS Algorithm}
\label{pseudo}
\begin{algorithmic}[1]
	\State $P^u \sim {\cal N}(0, I)$
    \State Select learning rates $\eta$ for the weights $\eta_w$ and label posteriors $\eta_{P^u}$
    \State \textbf{Phase I}: Estimate $P^u$
    \While{$P^u$ has not stabilized}
        \State $P^u = \Pi(P^u)$ (project posterior onto the probability simplex)
        \State $w_1 \sim {\cal N}(0, I)$
        \State $\Delta P^u = 0$
        \State // Run SGD for $T$ steps (on the weights) to estimate loss decrease
        \For{$t = 1 : T$}
        \State $w_{t-\frac{1}{2}} = w_{t-1} - \eta_{w} \nabla_{w_{t-1}} \left(\ell(B^u_t, P^u; w_{t-1})+ \beta q(B^u_t; w_{t-1}) \right)$
		\State $w_{t} = w_{t-\frac{1}{2}} - \eta_{w} \nabla_{w_{t-\frac{1}{2}}} \ell(B^l_t, P^l; w_{t-\frac{1}{2}})$
        \State $\Delta P^u = \Delta P^u + \nabla_{P^u} \ell(B^u_t, P^u; w_t)$
        \EndFor
        \State // Update the posterior distribution
        \State $P^u = P^u - \eta_{P^u}  \Delta P^u$ 
    \EndWhile
    \State \textbf{Phase II}: Estimate the weights.
    \State  $\hat y_i^u = \arg\max_i P^u_i \ \forall i=1,\ldots,N^u$ 
    \While {$w$ has not stabilized}
    \State  $w_1 \sim {\cal N}(0, I)$
     \State $w_{t-\frac{1}{2}} = w_{t-1} - \eta_{w} \nabla_{w_{t-1}} \frac{1}{|B^u_t|} \sum_{i = 1}^{|B^u_t|} \ell(x^u_i, \hat y^u_i; w_{t-1})$
     \State $w_{t} = w_{t-\frac{1}{2}} - \eta_{w} \nabla_{w_{t-\frac{1}{2}}} \frac{1}{|B^l_t|} \sum_{i = 1}^{|B^l_t|} \ell(x^l_i, y^l_i; w_{t-\frac{1}{2}})$
    \EndWhile
\end{algorithmic}
\end{footnotesize}
\end{algorithm}

It should be noted that the computation of the gradient $\nabla_{P^u} \ell(B^u_t, P^u; w_t)$ is not straightforward, as $w_t$ is, in general, a (complex) function of $P^u$. In the computation of the gradient, we omit here the dependence of $w_t$ on $P^u$, and use the approximation $\nabla_{P_i^u} \ell(w_t, x_i^u, P_i^u) \approx - \log f_{w_t} (x_i^u)$. This approximation is exact whenever each data point is visited once (i.e., $T = \text{1 epoch}$); as $T$ is chosen to be relatively small here, we assume that this approximation holds. 


It is important to note that, with the SaaS algorithm, we are not attempting to solve the optimization problem: $\min_{w, P^u} \sum_{i=1}^{N} \ell(x_i, P^u_i; w).$ This problem has many trivial solutions, as observed by \cite{zhang2016understanding}, as deep neural networks can easily fit random labels when trained long enough. Thus, for many posteriors $P^u$, there are weights $w$ achieving zero loss on this objective. One of many such trivial solutions is setting the label posterior $P^u$ to the outputs of the network trained only with the labeled samples. This would result in  the same test performance as that of a supervised baseline  and does not utilize the unlabeled samples at all. On the other hand, SaaS uses the cumulative loss up to a \textit{fixed}, small iteration $T$ as an inference criterion for label posterior $P^u$.

\begin{figure}
  \centering
    \includegraphics[width=6 cm]{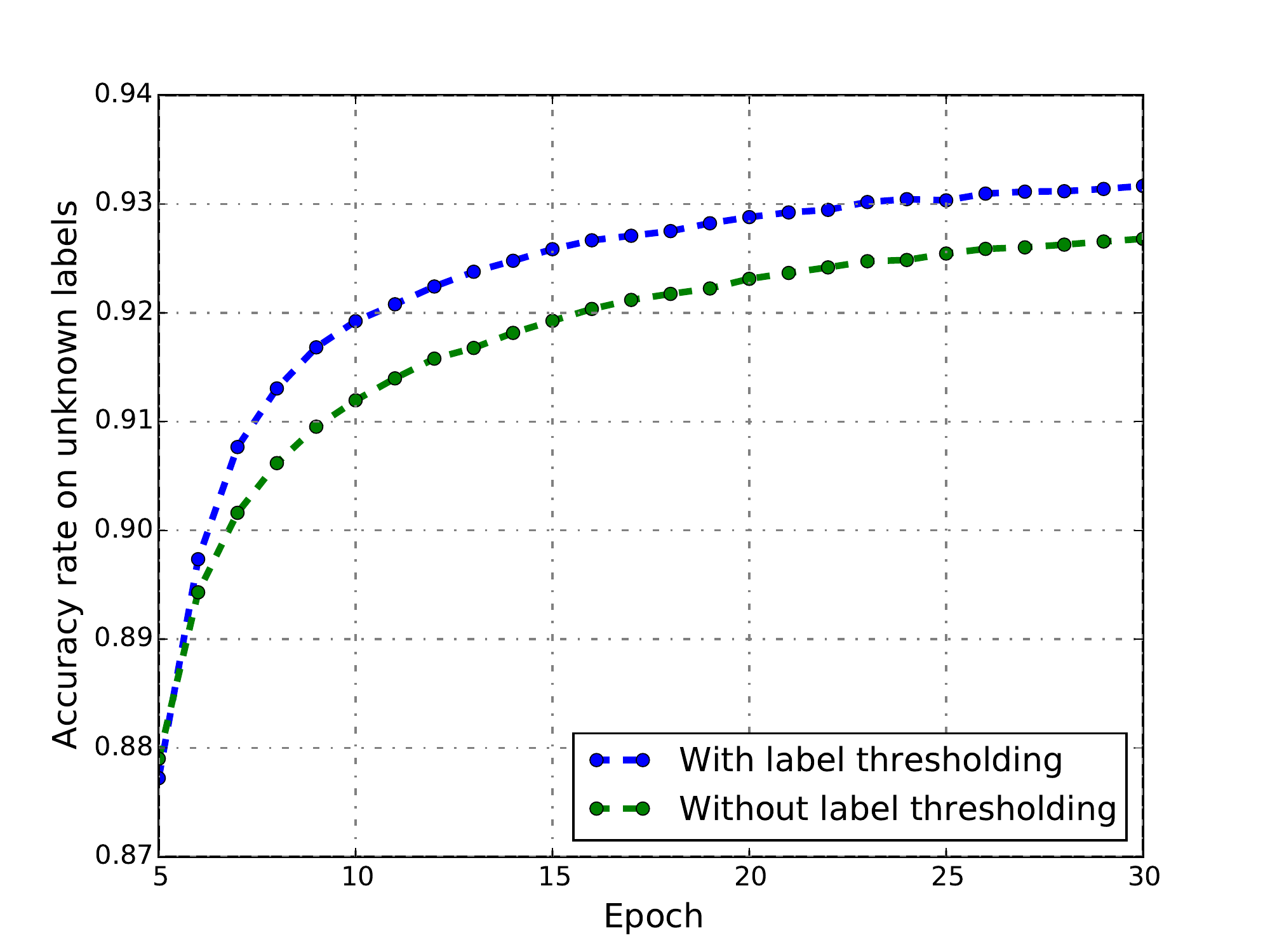}
    \includegraphics[width=6cm]{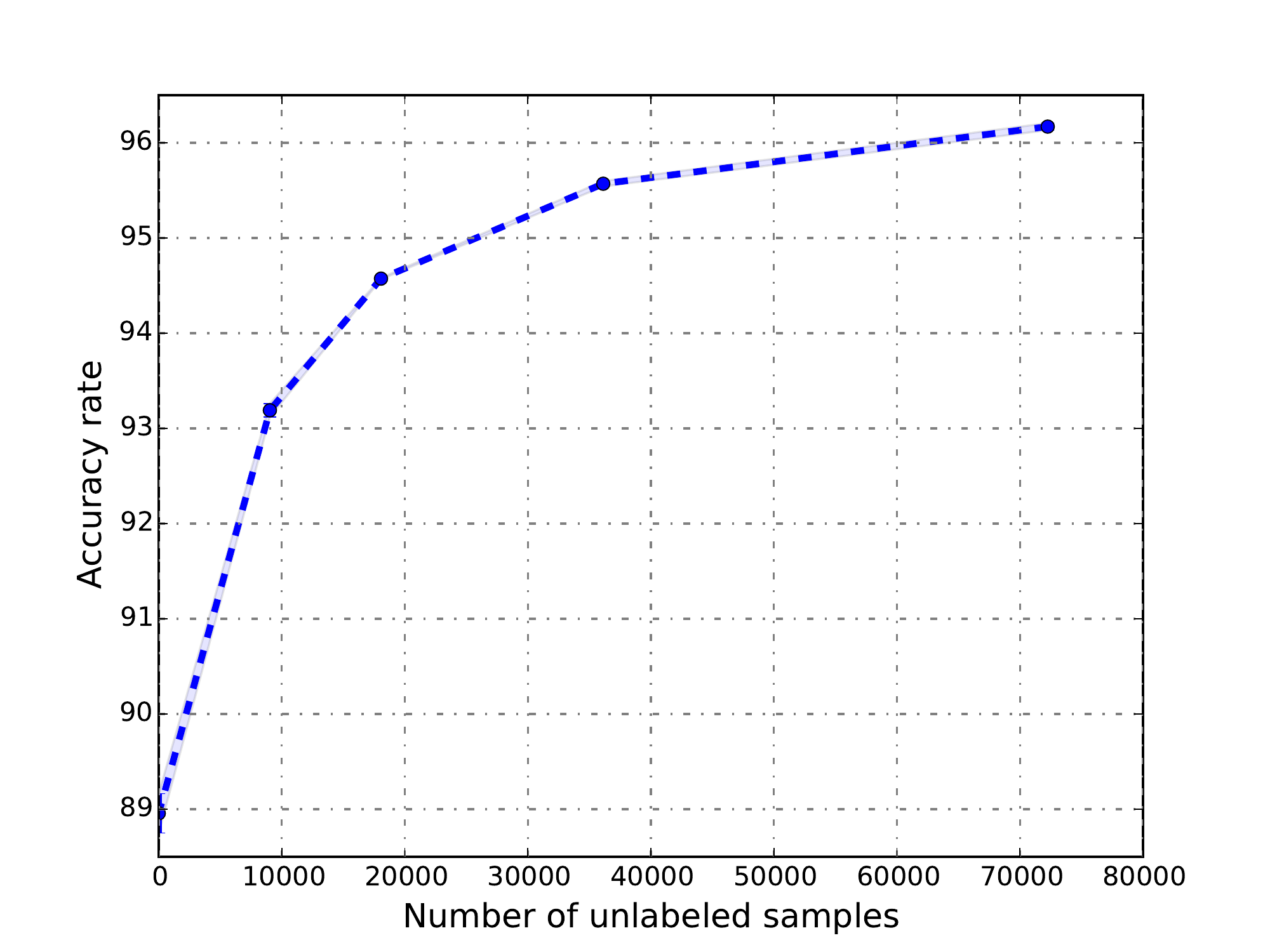}
\caption{(Left) {\bf Effect of label thresholding.} We project $P^u$ to the closest probability simplex with minimum probability for a class being $0.05$. Plot is given for unlabeled accuracy as label thresholding used in the first phase of the SaaS. The plot is given from epoch $5$ to epoch $30$. (Right) {\bf Test accuracy vs. number of unlabeled samples.} Accuracy on the test data versus number of unlabeled samples for the SVHN dataset using ResNet18. As the number of unlabeled samples increases, performance improves significantly, as expected in a semi-supervised learning scheme. Results are averaged over three random labeled sets, but error bars are not visible because deviations are smaller than the line-width.}
\label{thresholding}
\end{figure}

Finally, instead of projecting $P^u$ onto the probability simplex $\mathcal{S}$, we have found that the projection onto a slightly modified set $\mathcal{S}_\alpha = \{ x \in \mathbb{R}^K: \sum_i x_i = 1, x_i \geq \alpha \}$ (with $\alpha \geq 0$ chosen to be small) lead to better optimization results for $P^u$.  This is in line with recent work in supervised classification, where this technique is used in order to improve the accuracy of deep neural networks \cite{pereyra2017regularizing,szegedy2016rethinking}.  
Fig.~\ref{thresholding} (left) illustrates the effect of this approach for SaaS, and shows a clear improvement in SVHN dataset. 
\section{Empirical evaluation}
\label{sect-empirical}

We test the SaaS algorithm against the state-of-the-art in the most common benchmarks, described next.

\subsubsection*{Datasets.}

SVHN \cite{netzer2011reading} consists of images of house numbers. We use $73,257$ samples for training, rather than the entire $600,000$ images; $26,032$ images are separated for evaluation. CIFAR-10 \cite{krizhevsky2009learning} has $60,000$ images, of which $50,000$ are used for training and $10,000$ for testing. We choose labeled samples randomly. We also choose them to be uniform over the classes as it is done in previous works \cite{miyato2017virtual}. For both datasets, $10\%$ of the training set used for hyper-parameter tuning.

\subsubsection*{Training.}

As pointed out by \cite{zhang2016understanding}, deep networks can easily (over)fit random labels. We set $T$ small enough ($40$ epochs for CIFAR10 and $5$ epochs for SVHN) so that simulated weights cannot fit randomly initialized posterior estimates in the early epochs. We use ResNet18 \cite{he2016deep} as our architecture and vanilla SGD with momentum $0.9$ as an optimizer. We perform random affine transformations as data augmentations both in SVHN and CIFAR10. We additionally use horizontal flip and color jitter in CIFAR10. Learning rates for $w$ and $P^u$ are chosen as $\eta_w = 0.01$ and $\eta_{P^u} = 1$ respectively. We keep these rates fixed when learning $P^u$. We fixed the number of outer epochs as well for the first phase of the algorithm by setting it to $75$ for SVHN and $135$ for CIFAR10. For the second phase of SaaS (supervised part), training is not limited to a small epoch $T$. Instead, learning rate initialized as $0.1$ and halved after $50$ epochs unless accuracy in validation is increasing. We stop when the learning rate reaches $0.001$. 

The baseline for comparison is performance on the same datasets using only the labeled set (i.e. 4K samples for CIFAR10 and 1K samples for SVHN) (Fig.~\ref{results_sp_and_yu}). When training the (supervised) baseline, we employ the same learning parameters, architecture and augmentations as Phase II of SaaS. As expected, SaaS substantially improves baseline results, which is indicative that unlabeled data being effectively exploited by the algorithm. (Fig.~\ref{results_sp_and_yu})

In Fig.\ref{results_ssl}, we compare SaaS with state-of-the-art SSL methods on standard SSL benchmarks. In CIFAR-10, algorithms are trained with 4,000 labeled and 46,000 unlabeled samples. In SVHN, they are trained with 1,000 labeled and 72,257 unlabeled samples. The means and deviations of the test errors are reported by averaging over three random labeled sets. The state-of-the-art methods we compare include input smoothing algorithms \cite{miyato2017virtual}, ensembling models \cite{tarvainen2017mean,laine2016temporal}, generative models \cite{salimans2016improved} and models employing problem specific prior \cite{sajjadi2016regularization}. SaaS is comparable to state-of-the-art methods. Specifically, SaaS achieves the best performance in SVHN and second best result in CIFAR10 after VAT. Considering that VAT does input smoothing by adversarial training, our performance can be improved by combining with it. 

An SSL algorithm is expected to be more accurate when the number of unlabeled data increases. As it can be seen in Fig.\ref{thresholding} (right), we consistently get better results with more unlabeled samples. 

\begin{figure}
\begin{center}
 \begin{tabular}{||c | c | c||} 
 \hline
 Dataset & CIFAR10-4k & SVHN-1k \\ [0.5ex] 
 \hline\hline
 Error rate by supervised baseline on test data  & 17.64 $\pm$ 0.58 & 11.04 $\pm$  0.50\\
 \hline
 Error rate by SaaS on unlabeled data  & 12.81 $\pm$ 0.08 & 6.22 $\pm$ 0.02   \\
 \hline
 Error rate by SaaS on test data  & 10.94 $\pm$ 0.07 & 3.82$\pm$ 0.09\\
\hline
\end{tabular}
\caption{{\bf Baseline error rates.} Error rates on the benchmark test set for the baseline system trained with 4K labeled samples on CIFAR10 and 1K labeled samples on SVHN.  {\bf Error rate on unknown labels.} SaaS performance on the {\em unlabeled} set.  {\bf Error rate on test data.}  SaaS performance on the {\em test} set. Results are averaged over three random labeled sets. As it can be seen, results of SaaS on test data are significantly better than that of baseline supervised algorithm.}
\label{results_sp_and_yu}
\end{center}
\end{figure}

\begin{figure}
\begin{center}
 \begin{tabular}{||c | c | c||} 
 \hline
 Method-Dataset & CIFAR10-4k & SVHN-1k  \\ [0.5ex] 
 \hline\hline
 VAT+EntMin \cite{miyato2017virtual} & \textbf{10.55} &  3.86 \\ 
 \hline
 Stochastic Transformation \cite{sajjadi2016regularization} & 11.29 & NR \\
 \hline
 Temporal Ensemble \cite{laine2016temporal} & 12.16 & 4.42  \\
 \hline
 GAN+FM \cite{salimans2016improved} & 15.59 & 5.88  \\
  \hline
 Mean Teacher \cite{tarvainen2017mean} & 12.31 & 3.95  \\
 \hline
 SaaS   & 10.94 $\pm$ 0.07 & \textbf{3.82$\pm$ 0.09 } \\
 \hline
\end{tabular}
\caption{{\bf Comparison with the state-of-the-art.} Error rates on the test set are given for CIFAR10 and SVHN. NR stands for ``not reported.'' CIFAR10 is trained using 4K labels, SVHN using 1K. Results are averaged over three random labeled sets. Despite its simplicity, SaaS performs at the state-of-the-art. It could be combined with adversarial examples (VAT) but here we report the naked results to highlight the role of speed as a proxy for learning in a semi-supervised setting while maintaining a simple learning scheme.}
\label{results_ssl}
\end{center}
\end{figure}

\begin{figure}
  \centering
    \includegraphics[width=6cm]{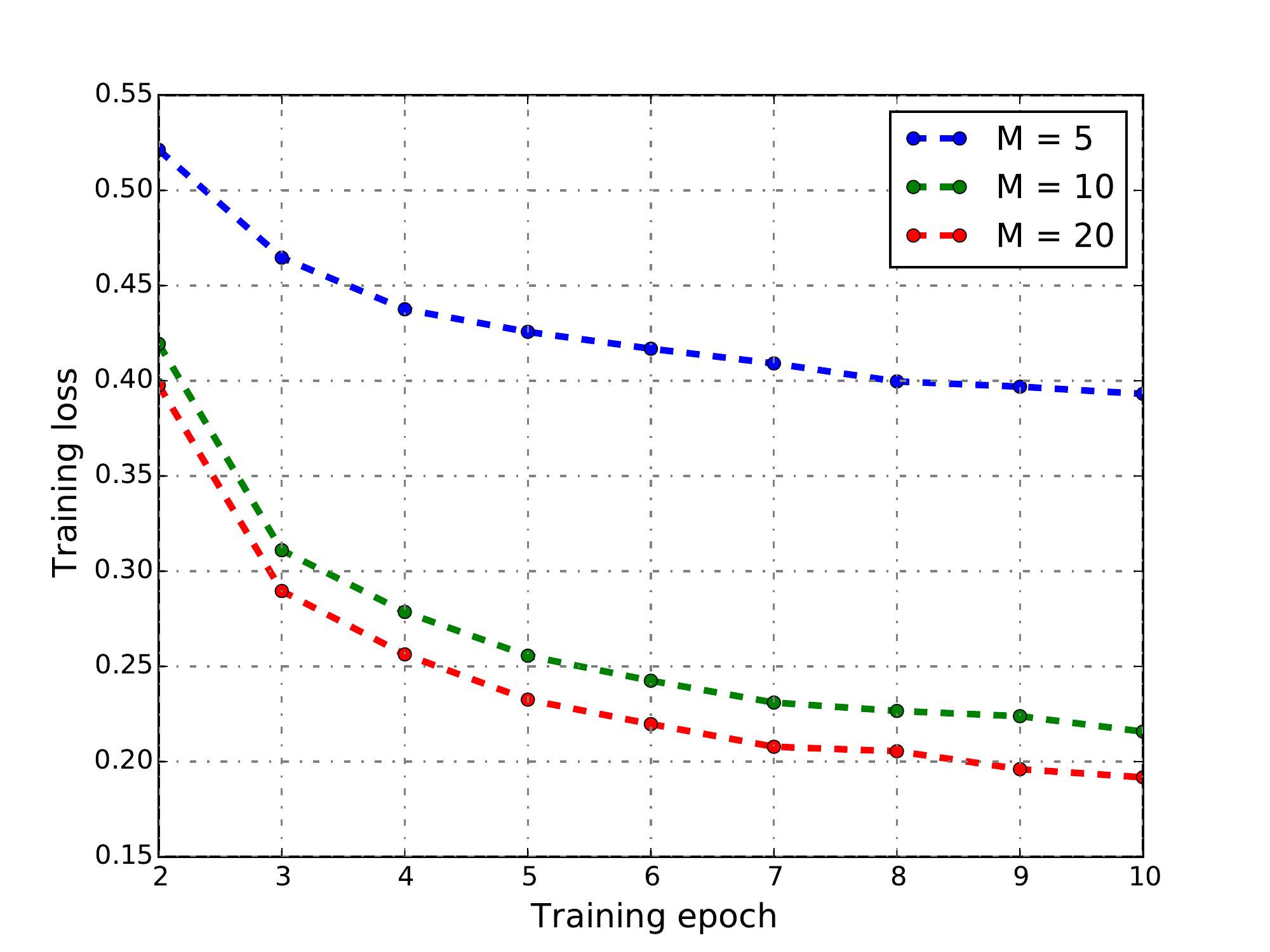}
    \includegraphics[width=6cm]{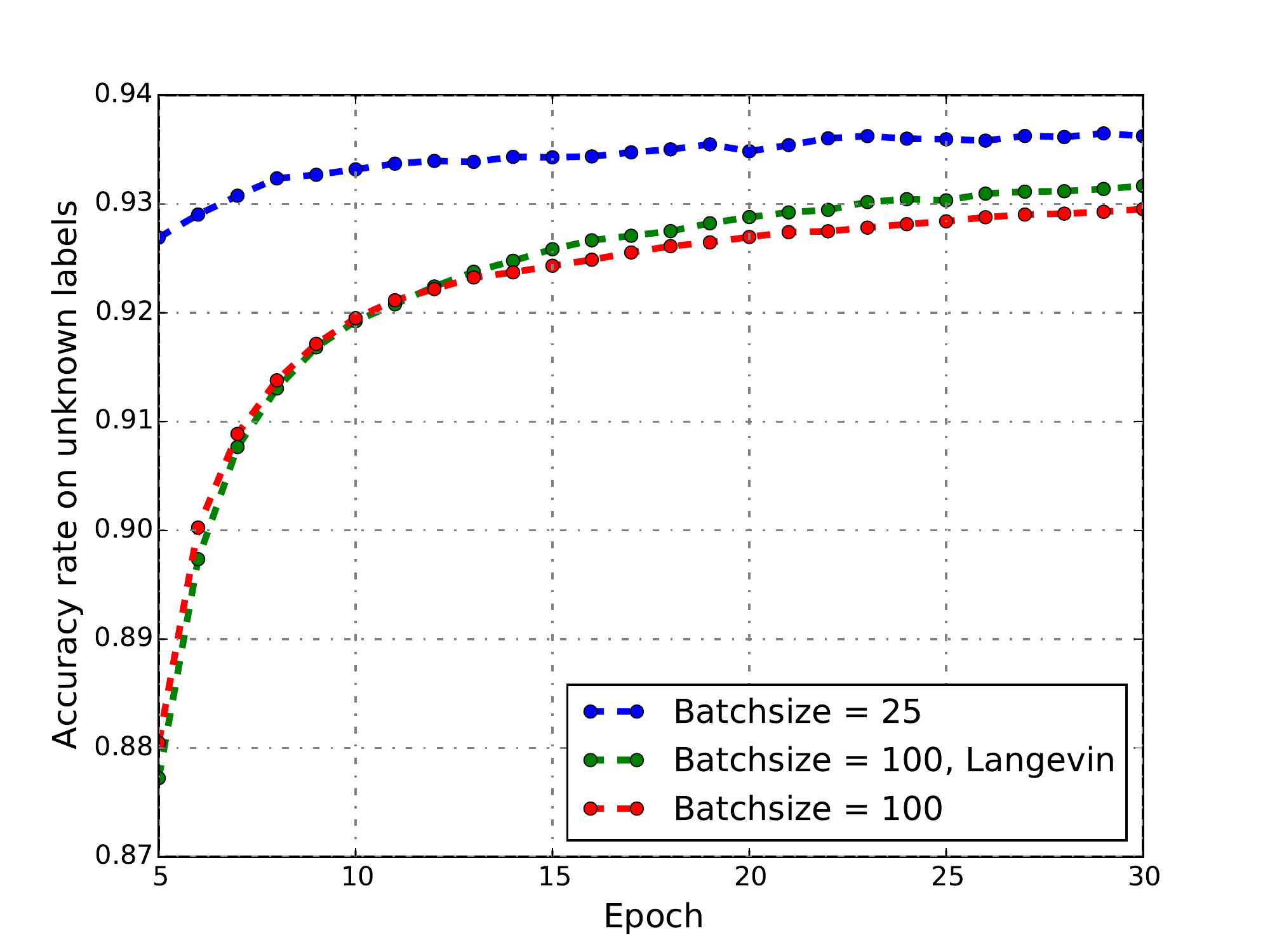}
\caption{(Left) {\bf SaaS finds pseudo-labels on which training is faster.} Training loss for a network trained with given $P^u$, as estimated by the first phase of our algorithm with different numbers of outer epochs $M$. $M$ is the number of epoch at which outer iteration stopped in the SaaS. Label hypotheses generated by our algorithm lead to faster training as the iteration count increases. Losses are plotted starting from epoch $2$. This plot verifies that our algorithm finds labels on which training is faster. (Right) {\bf Effect of Langevin.} Performance improves with smaller batch size, albeit at a significant computational cost: algorithm is about three times slower for $|B|=25$ (plots shown for SVHN). We therefore choose  $|B| = 100$ and add zero-mean Gaussian noise to the weight updates for comparable results (Langevin). The results converge to those of $|B|=25$ when we train longer. Plot is given for unlabeled accuracy because Langevin is only used in the first phase of SaaS. The plot is given from epoch $5$ to epoch $30$.}  
\label{langevin}
\end{figure}

We motivated SaaS as a method finding labels for which training decrease in a fix small number of epochs (e.g. $10$) is the maximum. To verify that our algorithm actually does what is intended to do, we train networks on the pseudo-labels generated by SaaS. One can see in Fig.\ref{langevin} (left) that as SaaS iterates more (i.e. as the number of updates for $P^u$ increases), resulting pseudo-labels leads to larger training loss decrease (faster training) in the early epochs. This experiment verifies that SaaS gives pseudo-labels on which training would be faster.

The results reported in Fig.~\ref{results_sp_and_yu} are with ResNet18 and affine augmentations. Our method uses augmentation, but for direct comparison with some of the previous papers, we also report results with the convolutional network ``conv-large'' and translational augmentations as used in \cite{miyato2017virtual,tarvainen2017mean} in Fig.~\ref{results_CL}. Additionally, horizontal flipping is used in CIFAR10. Moreover, we applied pre-processing by centering relative to the Mahalanobis metric (known as ZCA) as in \cite{miyato2017virtual,tarvainen2017mean}. 

\begin{figure}
\begin{center}
\begin{tabular}{||c | c | c||} 
\hline
Dataset & CIFAR10-4k & SVHN-1k \\ [0.5ex] 
\hline\hline
Error rate by supervised baseline on test data & 17.88 $\pm$ 0.19 & 12.72 $\pm$ 1.13 \\
\hline
Error rate by SaaS on unlabeled data & 14.26  $\pm$ 0.30 & 7.26 $\pm$ 0.19  \\
\hline
Error rate by SaaS on test data  & 13.22 $\pm$ 0.31 &  4.77 $\pm$ 0.27\\
\hline
\end{tabular}   
\caption{For direct comparison, we implement SaaS with the ``conv-large'' architecture of \cite{miyato2017virtual} and the same augmentation scheme. Baseline performance (supervised) is also shown. SaaS improves both on the unlabeled set and test set. Results are averaged over three random labeled sets.}
\label{results_CL}  
\end{center}
\end{figure}

\subsubsection*{Small batch-size and Langevin dynamics.}

Finally, we discuss a method we use to reduce the training time for SaaS. We achieve better performance with smaller batch size $|B| = 25$ for both labeled and unlabeled data. When  $|B| = 100$, generalization performance degrades as expected \cite{keskar2016large}. Unfortunately, small batch-size slows down training, so we use $|B| = 100$ for both labeled and unlabeled data and add zero-mean Gaussian noise to the weight updates, a process known as stochastic gradient Langevin dynamics (SGLD) \cite{welling2011bayesian,raginsky2017non,chaudhari2016entropy}, with variance  $10^{-5}\eta_w$ for all the datasets. Comparison of small and large batches without noise and large batches with noise can be seen in Fig.~\ref{langevin} (right). With this, first phase of the algorithm (getting the estimates of unknown labels) takes about $1$ day for SVHN and $4$ days for CIFAR10 using GeForce GTX $1080$ when we use ResNet18. 

\section{Discussion and related work}
\label{sect-discussion}

The key idea of our approach to SSL is to leverage on training speed as a proxy to measure the quality of putative labels as they are iteratively refined in a differentiable manner.

That speed of convergence relates to generalization is implicit in the work of \cite{hardt2015train}, who derive an upper bound on generalization error as a function of a constant times the sum of step sizes, suggesting that faster training correlates with better generalization. 

Another way of understanding our method is via shooting algorithms used to solve boundary value problems (BVP). In a BVP with second order dynamics, a trajectory is found by simulating it with a guess of initial state; then, the initial state is refined iteratively such that the target error would be minimized. In our problem, dynamics are given by SGD. Assuming that we use SGD without momentum, we have a first order differential equation. The first boundary condition is the initialization of weights and the second boundary condition is a small cumulative loss. The latter one is used to refine $P^u$ which is a parameter of the dynamics instead of the initial state.   

In the next paragraphs we discuss our contribution in relation to the vast and growing literature on SSL.  

\subsubsection*{Ensemble Methods} include teacher-student models, that use a combination of estimates (or weights) of classifiers trained under stochastic transformations. Although we train only one network, our method resembles the teachers-student models: Our $P^u$ update is similar to a teacher classifier in the teacher-student models. However, we randomly start a student model at each outer epoch. In \cite{laine2016temporal} the prediction of the network over the training epochs are averaged, whereby in each epoch a different augmentation is applied. \cite{tarvainen2017mean} minimize the consistency cost, which is the distance between two network outputs. Hence, the student network minimizes classification and consistency costs with labeled data and only consistency with the unlabeled data. The weights of the teacher model are the running average of the weights of the student network. 

\subsubsection*{Cluster assumption.}
The cluster assumption posits that inputs with the same class are in the same cluster under an appropriate metric. It takes many forms (max-margin, low-density separation, smoothness, manifold). In general, it could be framed as $\int || \nabla_x f_w(x) ||^2 d \mu_x$ being small where $\mu_x$ is the probability distribution over some manifold. VAT \cite{miyato2015distributional,miyato2017virtual} is a recent application of this idea to deep networks, realized by adding a regularization term to minimize the difference between the network outputs for clean and adversarial noise-added-inputs. This state-of-the-art method is similar to the adversarial training of \cite{goodfellow2014explaining}, the main difference being that it does not require label information, and thus can be applied to SSL. Our method is orthogonal to VAT and can be improved by combining with it.

\subsubsection*{Self-training}
is an iterative process where confident labels from previous iterations are used as ground truth. In \cite{blum1998combining}, disjoint subsets of features of labeled samples are used to produce different hypotheses on randomly selected subsets of the unlabeled data. Labeled data are extended with the most confident estimates on this subset. This approach fails to enlarge the sigma-algebra generated by the labeled samples and generally fails if the classifier does not give correct estimates for at least one feature subset. We maintain an estimate of the posterior probability of each label, and only force a point estimate in the refinement (second) phase of the algorithm.

\subsubsection*{Encoding priors.}
In image classification one can enforce invariance of labels to some transformations. This is achieved by minimizing the difference between network outputs under different transformations. In \cite{sajjadi2016regularization}, transformations are affine (translation, rotation, flipping, stretching and shearing). Although they achieve good results, their improvement on {baseline} supervised performance (using only labeled data) is marginal. {\em E.g.}, in CIFAR-10 supervised error is $13.6\%$ while semi-supervised error is $11.29\%$. Similarly, \cite{simard1992tangent} suggests minimizing the norm of directional derivatives of the network with respect to small transformations. We also employ augmentations like most SSL papers on image classification. 

\subsubsection*{Generative models} used to be the standard for SSL, but the high dimensionality of problems in vision presents a challenge. Adversarial methods like GANs have been recently applied, whereby  an additional $C+1$-th (fake) class is used. The loss function is designed to force the discriminator output to be low for the fake class for the unlabeled samples while making it high for the generated samples. \cite{salimans2016improved} suggested a regularizer for the generator, called feature matching (FM), whereby the generator tries to match the first-order statistics of the generated sample features to those of the real data. According to \cite{dai2017good}, the discriminator benefits the generator if it has samples within the data manifold, but around subspaces in which the density of samples is low. Unlike feature matching, they match the inverse distribution in non-zero density areas rather than their means. Instead of one generator network, \cite{dumoulin2016adversarially} uses an encoder-decoder network generating images and labels from which the discriminator tries to differentiate.  

\subsubsection*{Graph based methods.} \cite{yang2016revisiting} assumes that an affinity matrix of size $N \times N$ is given, which has information independent from the one in the features of the data. In the loss function, they have a term penalizing  different labels assigned to similar samples based on this similarity matrix. \cite{nie2011unsupervised,su2016efficient} finds a sparse clustering using the $\ell_1$-norm. \cite{lu2010constrained} propagates pairwise \textit{must} and \textit{cannot} constraints in an efficient way. \cite{li2015learning} uses the current hypothesis for the unknown labels in learning as in our algorithm. They update the affinity matrix and estimates of the unknown labels iteratively. \cite{wang2015adaptively} suggests a dictionary learning method which can be used for SSL. Recent graph based methods \cite{haeusser2017learning,gaunt2018graph,kipf2016semi,weston2012deep} exploit deep networks for function approximation in a manner that can be used for SSL.

Within this rich and multi-faceted context, our approach provides one more element to consider: The fact that the speed of convergence when optimizing with respect to the probability of unknown labels is highly dependent on their correctness, even when starting from a random initial condition. This frees us from having to jointly optimized the parameters and the posterior on the labels, which would blow up the dimensionality, and allows us to focus sequentially on first estimating the unknown label distribution -- irrespective of the model parameters/weight -- and then retrieve the weights using the maximum a-posteriori estimate of the labels.

Our method can be combined with other ideas recently introduced in SSL, including using adversarial examples. We do not do so in our experiments, to isolate the contribution of our algorithm. Nevertheless, just the method alone, with some data augmentation but without sophisticated tricks, achieves state-of-the-art performance.

Our method requires some data augmentation. While even translation and horizontal flipping suffice, performance suffers in the absence of any augmentation. More experiments are provided in the Supplementary Material.

An implementation of our algorithm will be made public in source format upon completion of the anonymous review process. 

\bibliographystyle{splncs}
\bibliography{ref.bib}

\end{document}